\documentclass[runningheads]{llncs}

\usepackage[year=2024,ID=8392]{eccv}
\usepackage{eccvabbrv}

\usepackage{color,xcolor}
\usepackage{epsfig}
\usepackage{graphicx}

\usepackage{adjustbox}
\usepackage{array}
\usepackage{booktabs}
\usepackage{colortbl}
\usepackage{float,wrapfig}
\usepackage{hhline}
\usepackage{multirow}
\usepackage{subcaption} % issues a warning with CVPR/ICCV format
\usepackage[toc,page]{appendix}

\usepackage{amsmath,amsfonts,amssymb}
\usepackage{bm}
\usepackage{nicefrac}
\usepackage{microtype}
\usepackage{inconsolata}
\usepackage{pifont}

\usepackage{changepage}
\usepackage{extramarks}
\usepackage{fancyhdr}
\usepackage{lastpage}
\usepackage{setspace}
\usepackage{soul}
\usepackage{xspace}
\usepackage{indentfirst}
\usepackage{paralist}
\usepackage{booktabs,arydshln}

\usepackage[breaklinks,colorlinks,citecolor=eccvblue]{hyperref}
\usepackage{orcidlink}
\usepackage{url}
\usepackage[frozencache,cachedir=minted]{minted}

\usepackage{algorithm, algorithmic}
\usepackage{enumerate}
\usepackage{lipsum}
\usepackage{tikz}
\usetikzlibrary{tikzmark}

\newcolumntype{L}[1]{>{\raggedright\let\newline\\\arraybackslash\hspace{0pt}}m{#1}}
\newcolumntype{C}[1]{>{\centering\let\newline\\\arraybackslash\hspace{0pt}}m{#1}}
\newcolumntype{R}[1]{>{\raggedleft\let\newline\\\arraybackslash\hspace{0pt}}m{#1}}

\newcommand{\sect}[1]{Section~\ref{#1}}

\newcommand{\fig}[1]{Figure~\ref{#1}}

\newcommand{\tab}[1]{Table~\ref{#1}}

\newcommand{\ignorethis}[1]{}

\makeatletter
\DeclareRobustCommand\onedot{\futurelet\@let@token\@onedot}
\def\@onedot{\ifx\@let@token.\else.\null\fi\xspace}

\def\eg{\emph{e.g}\onedot} 
\def\ie{\emph{i.e}\onedot}

\makeatother

\makeatletter
\def\adl@drawiv#1#2#3{%
        \hskip.5\tabcolsep
        \xleaders#3{#2.5\@tempdimb #1{1}#2.5\@tempdimb}%
                #2\z@ plus1fil minus1fil\relax
        \hskip.5\tabcolsep}
\newcommand{\cdashlinelr}[1]{%
  \noalign{\vskip\aboverulesep
           \global\let\@dashdrawstore\adl@draw
           \global\let\adl@draw\adl@drawiv}
  \cdashline{#1}
  \noalign{\global\let\adl@draw\@dashdrawstore
           \vskip\belowrulesep}}
\makeatother

\newcommand{\cmark}{\ding{51}}
\newcommand{\xmark}{\ding{55}}

\definecolor{mydarkblue}{rgb}{0,0.08,1}
\definecolor{mydarkgreen}{rgb}{0.02,0.6,0.02}
\definecolor{darkred}{rgb}{0.8,0.02,0.02}
\definecolor{darkorange}{rgb}{0.40,0.2,0.02}
\definecolor{darkpurple}{RGB}{111,0,255}
\definecolor{myred}{rgb}{1.0,0.0,0.0}
\definecolor{mygold}{rgb}{0.75,0.6,0.12}
\definecolor{mydarkgray}{rgb}{0.66, 0.66, 0.66}

\newcommand{\myparagraph}[1]{\vspace{-8pt}\paragraph{#1}}

\usepackage[accsupp]{axessibility}
\usepackage{listings}

\begin{document}

\title{Sparse Refinement for Efficient High-Resolution Semantic Segmentation} 
\titlerunning{Sparse Refinement for Efficient High-Resolution Semantic Segmentation}
\author{
Zhijian Liu\textsuperscript{1,2,$*$} \hspace{5mm} Zhuoyang Zhang\textsuperscript{3,$*$} \hspace{5mm} Samir Khaki\textsuperscript{4} \hspace{5mm} Shang Yang\textsuperscript{1} \\
Haotian Tang\textsuperscript{1} \hspace{5mm} Chenfeng Xu\textsuperscript{5} \hspace{5mm} Kurt Keutzer\textsuperscript{5} \hspace{5mm} Song Han\textsuperscript{1,2}
}
\authorrunning{Liu\textsuperscript{$*$}, Zhang\textsuperscript{$*$}, Khaki, Yang, Tang, Xu, Keutzer, and Han}
\institute{\textsuperscript{1}MIT \hspace{2mm} \textsuperscript{2}NVIDIA \hspace{2mm} \textsuperscript{3}Tsinghua University \hspace{2mm} \textsuperscript{4}University of Toronto \hspace{2mm} \textsuperscript{5}UC Berkeley\\
\url{https://sparserefine.mit.edu}}

\maketitle

\footnotetext{$*$ indicates equal contributions.}

\begin{abstract}

Semantic segmentation empowers numerous real-world applications, such as autonomous driving and augmented/mixed reality. These applications often operate on high-resolution images (\eg, 8 megapixels) to capture the fine details. However, this comes at the cost of considerable computational complexity, hindering the deployment in latency-sensitive scenarios. In this paper, we introduce \textbf{SparseRefine}, a novel approach that enhances \textit{dense low-resolution} predictions with \textit{sparse high-resolution} refinements. Based on coarse low-resolution outputs, SparseRefine first uses an entropy selector to identify a sparse set of pixels with high entropy. It then employs a \textit{sparse} feature extractor to efficiently generate the refinements for those pixels of interest. Finally, it leverages a gated ensembler to apply these sparse refinements to the initial coarse predictions. SparseRefine can be seamlessly integrated into any existing semantic segmentation model, regardless of CNN- or ViT-based. SparseRefine achieves significant speedup: \textbf{1.5 to 3.7 times} when applied to HRNet-W48, SegFormer-B5, Mask2Former-T/L and SegNeXt-L on Cityscapes, with negligible to no loss of accuracy. Our ``\textit{dense+sparse}'' paradigm paves the way for efficient high-resolution visual computing.

\end{abstract}
\section{Introduction}

Semantic segmentation is a fundamental computer vision task with critical applications in autonomous driving, augmented reality, and mixed reality. Deep neural networks have significantly boosted semantic segmentation performance in recent years~\cite{long2015fully,chen2017rethinking,wang2020deep,xie2021segformer,cheng2022masked,guo2022segnext}. Yet, deploying these computationally intensive models on resource-constrained edge devices remains a challenge.

Significant efforts have been dedicated to designing compact neural networks with reduced computational complexity~\cite{howard2017mobilenets,sandler2018mobilenetv2,ma2018shufflenet,zhang2018shufflenet,iandola2016squeezenet}. However, in dense-prediction tasks like semantic segmentation, the image resolution makes greater impact to model's inference latency than the model size. This is because real-world segmentation applications often involve megapixel high-resolution images, which surpass the typical image classification workload by 1-2 orders of magnitude.

Reducing the image resolution through downsampling can result in a noticeable increase in speed. But, this comes at the cost of accuracy degradation. Segmentation models are generally more adversely affected by reduced resolution compared to classification models as low-resolution images result in the loss of fine details, including small or distant objects. The missing information can be safety-critical (\eg, for autonomous driving). 

\begin{figure}[t]
\centering
\includegraphics[width=\linewidth]{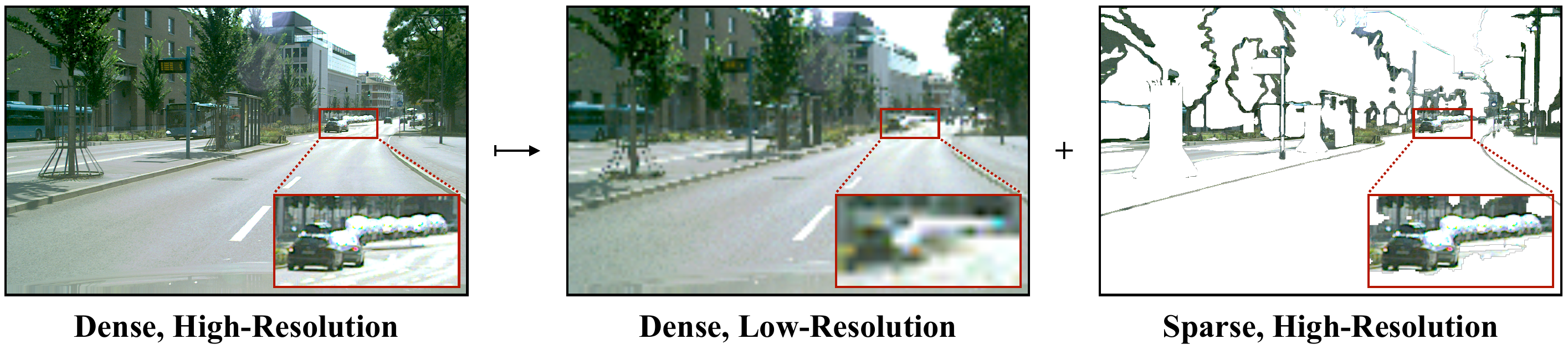}
\caption{Processing \textit{dense high-resolution} inputs is computationally expensive. In this paper, we propose an alternative approach by integrating \textit{dense low-resolution} and \textit{sparse high-resolution} inputs, which provide complementary information about the overall scene layout and intricate object details. Leveraging the lower resolution and sparsity of these inputs allows for more efficient processing.}
\label{fig:teaser}
\vspace{-12pt}
\end{figure}

This paper introduces \textbf{SparseRefine} as a novel and complementary approach to address this problem. We find that the differences between the dense low-resolution predictions and the dense high-resolution predictions primarily emerge in a sparse set of pixels.
As shown in Figure~\ref{fig:teaser}, our idea is to enhance \textit{dense low-resolution} predictions (based on downsampled inputs) with \textit{sparse high-resolution} refinements. SparseRefine only refines a sparse set of carefully-selected pixels, enabling it to avoid unnecessary high-resolution computations at the fine-grained pixel level. Besides, SparseRefine is compatible with both CNN- and ViT-based semantic segmentation models.

SparseRefine achieves remarkable and consistent speedup: \textbf{1.5 to 3.7 times} when applied to HRNet-W48, SegFormer-B5, Mask2Former-T/L and SegNeXt-L on Cityscapes, while maintaining accuracy. We also validate the general effectiveness of SparseRefine on many other datasets including Pascal VOC~\cite{everingham2010pascal}, BDD100K~\cite{yu2020bdd100k}, Deepglobe~\cite{demir2018deepglobe}, and ISIC~\cite{codella2018skin}. SparseRefine also exhibits superior performance compared with related methods including Token Pruning~\cite{chen2023sparsevit}, Mask Refinement~\cite{kirillov2020pointrend}, and Patch Refinement~\cite{huang2019uncertainty, verelst2022segblocks, wu2020patch}.

\section{Related Work}

\subsection{Semantic Segmentation} Semantic segmentation is a fundamental task in computer vision which assigns a class label to each pixel in an image. Following FCN~\cite{long2015fully}, early deep learning models~\cite{badrinarayanan2015segnet,ronneberger2015u} for semantic segmentation relied on CNN-based architectures. DeepLab and PSPNet~\cite{zhao2017pyramid} improved FCN by introducing atrous convolution~\cite{chen2014semantic}, spatial pyramid pooling~\cite{he2014spatial,zhao2017pyramid,chen2016deeplab}, encoder-decoder mechanism~\cite{chen2017deeplabv3}, depthwise convolution~\cite{chen2018deeplabv3plus} and neural architecture search~\cite{liu2019autodeeplab}. Follow-up research proposed attention mechanism~\cite{fu2019dual} and object context modeling~\cite{yuano2020bject}. Recently, researchers also studied efficient segmentation architectures~\cite{paszke2016enet,wu2019fastfcn,poudel2019fast,mehta2018espnet,yu2018bisenet,yu2021bisenet,zhao2018icnet,chen2019fasterseg,hong2021deep,guo2022segnext}.
Recent advances in vision transformers~\cite{dosovitskiy2020image,liu2021swin,liu2022swin,wang2021pyramid,yuan2021tokens,fan2021multiscale} also inspired the design of attention-based semantic segmentation models. SegFormer~\cite{xie2021segformer}, SETR~\cite{zheng2021rethinking}, Segmenter~\cite{strudel2021segmenter}, HRFormer~\cite{yuan2021hrformer}, SwinUNet~\cite{cao2022swin} and EfficientViT~\cite{cai2022efficientvit} designed transformer-based backbones for segmentation, while MaskFormer~\cite{cheng2021maskformer} and Mask2Former~\cite{cheng2022masked} modeled semantic segmentation as mask classification. 

\subsection{Activation Sparsity} Activation sparsity naturally exists in videos~\cite{pan2018recurrent,pan2021va}, point clouds~\cite{graham20183d,liu2022spatial} and masked images in self-supervised visual pre-training~\cite{he2022masked,gao2022convmae,tian2023designing,huang2022green}. It can also be introduced through activation pruning~\cite{pan2021iared2,rao2021dynamicvit,kong2022svit,yin2021adavit,song2022cpvit,liang2022not}, token merging~\cite{bolya2022tome,bolya2023tomesd} or clustering~\cite{ma2023image}. These methods are specifically designed for classification or detection tasks, where there is no need to preserve information from all pixels. However, they are not suitable for semantic segmentation, which requires per-pixel predictions. An exception is SparseViT~\cite{chen2023sparsevit}, which skips computation on pruned windows while retaining their features. As such, SparseViT also works for semantic segmentation tasks. We will demonstrate that SparseRefine achieves superior efficiency compared with SparseViT in \sect{sect:exp}. Recently, system and architecture researchers also created high-performance GPU libraries~\cite{ren2018sbnet,choy20194d,yan2018second,tang2022torchsparse,tang2023torchsparsev2,hong2023pcengine} and specialized hardware~\cite{zhang2020sparch,wang2021spatten,lin2021pointacc,gondimalla2019sparten,wang2021dual} to exploit activation sparsity.

\subsection{Mask Refinement} Mask refinement for segmentation has been studied even before the prevalence of deep learning. Traditional methods~\cite{shi00normalized,boykov2001fast,felzenszwalb2004efficient,rother2004grabcut} formulated the task of semantic segmentation as graph cuts. The mask outputs were then post-processed using a conditional random field (CRF)~\cite{lafferty2001conditional,blake2004interactive,krahenbuhl2011efficient}, which aimed to minimize energy and capture local consistency in predicted labels. While CRF continues to impact the field in the deep learning era~\cite{chen2016deeplab,wu2018squeezeseg,choy20194d}, its inefficiency eventually led to the development of PointRend~\cite{kirillov2020pointrend} and RefineMask~\cite{zhang2021refinemask}. Inspired by graphics rendering, PointRend~\cite{kirillov2020pointrend} first identifies uncertain pixels from deeper and lower resolution feature maps. These pixels are then refined using a PointNet~\cite{qi2017pointnet}, leveraging interpolated shallower and higher resolution features. RefineMask~\cite{zhang2021refinemask} gradually upsamples the predictions and incorporates the fine-grained features to alleviate the loss of details for high-quality instance mask prediction. Both PointRend and RefineMask upscale the \textit{output resolution} with the help of high-resolution \textit{features}, while SparseRefine is focused on reducing the \textit{input resolution} and retains fine-grained details from full-scale \textit{raw RGB pixels}. While PointRend and RefineMask prioritize improving \textit{accuracy}, SparseRefine aims to minimize \textit{latency}. Therefore, our method is fundamentally orthogonal to existing mask refinement strategies.

\subsection{Multi-Scale Models} Multi-scale models have garnered popularity in high-resolution visual recognition tasks due to the diverse range of object sizes within an image. In early segmentation approaches, multi-resolution features were fused either using an FPN~\cite{lin2017refinenet,kirillov2019panoptic,yu2018bisenet,yu2021bisenet} or right before the prediction head~\cite{chen2016deeplab,chen2017deeplabv3,he2014spatial}. Subsequently, new primitives such as OctaveConv~\cite{chen2019drop}, HRNet~\cite{wang2020deep}, and DDRNet~\cite{wang2021dual} were designed to more effectively leverage multi-scale features within the backbone. There have also been explorations on refining the predictions in a patch-wise manner~\cite{verelst2022segblocks,wu2020patch,huang2019uncertainty}. Unlike SparseRefine, which enhances dense low-resolution predictions with \textit{sparse} high-resolution details, existing methods focus on performing \textit{dense} refinements. Also, while existing multi-scale models employ a \textit{parallel} design for their low-resolution and high-resolution modules, SparseRefine adopts a \textit{sequential} counterpart. This makes our method orthogonal to these designs. We will show in \sect{sect:exp} that SparseRefine could bring further improvements to multi-scale models (\eg, HRNet). 
\section{SparseRefine}

\begin{figure}[t]
\centering
\includegraphics[width=\linewidth]{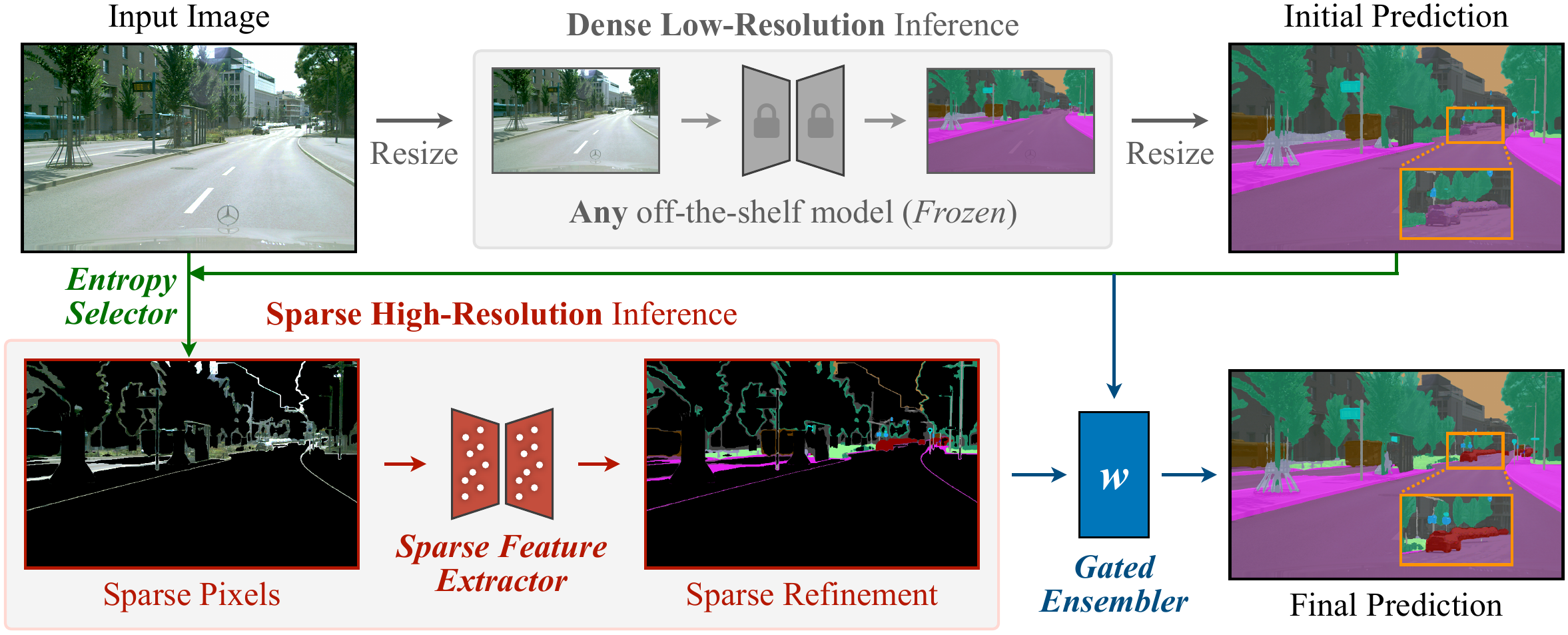}
\caption{SparseRefine improves initial \textit{dense low-resolution} predictions with \textit{sparse high-resolution} refinements. It first performs the dense low-resolution inference on the downsampled image to obtain the initial prediction. Subsequently, it uses an entropy selector to identify a sparse set of pixels with high entropy, and then employs a \textit{sparse} feature extractor to efficiently generate refinements for those selected pixels. Afterwards, it applies these sparse refinements to the initial predictions with a gated ensembler.}
\label{fig:method:overview}
\vspace{-12pt}
\end{figure}

SparseRefine improves dense low-resolution predictions with sparse high-resolution refinements. Figure \ref{fig:method:overview} provides an overview of our pipeline. It first downsamples the input image and performs the dense low-resolution inference to obtain the initial prediction. Then, it uses an entropy selector to identify a sparse set of pixels from the input image that exhibit high entropy in the initial prediction. High-entropy pixels demonstrate high prediction uncertainty and are likely misclassified ones. Subsequently, a sparse feature extractor is employed on the selected pixels to efficiently generate refinements. The sparse feature extractor operates on high-resolution pixels, allowing it to capture fine-grained details that may be overlooked in the dense low-resolution inference. Finally, these sparse refinements are applied to the initial predictions using a gated ensembler to obtain the final prediction. Training and inference share the same pipeline, where the model used in the dense low-resolution inference is trained in advance and keeps frozen during the SparseRefine training process.

\subsection{Dense Low-Resolution Prediction}
\label{sec:dense-low}

One of the most straightforward ways for accelerating inference is downsampling the input image. For instance, halving the resolution of HRNet-W48~\cite{wang2020deep} will lead to a 3.7$\times$ speedup, close to the theoretical computation reduction of 4$\times$. To enable inference on downsampled images, we employ an off-the-shelf segmentation model and train it on downsampled images. Since semantic segmentation models inherently support varying image resolutions, no modifications to the model architecture are necessary. For clarity, we will refer to the model trained here as the dense baseline model.

We first obtain coarse predictions from the downsampled images. Our refinement process then proceeds independently of the original dense segmentation model. This makes our refinement module an \textit{add-on} that can seamlessly enhance any off-the-shelf model. Next, we upsample the coarse predictions (using nearest neighbor interpolation) to match the original input resolution. All subsequent refinements will be built upon this. 

The downsampling process inevitably leads to information loss, causing a decline in accuracy.  In the following section, we will demonstrate how our method effectively addresses this accuracy gap through efficient sparse refinement on selected high-resolution pixels.

\subsection{Sparse High-Resolution Refinement}

Low-resolution predictions are fast but not as accurate as high-resolution predictions. Fortunately, \textit{the differences in their predictions primarily emerge in a sparse set of pixels}, often associated with small or distant objects and object boundaries. Building upon this observation, our objective is to \textit{sparsely} refine the less accurate predictions so that we could bridge the accuracy gap efficiently.

\subsubsection{Entropy Selector.} The selection of sparse pixels plays a critical role in our entire pipeline as it directly determines the number and specific pixels on which we apply the refinement process. Ideally, we would want to choose those pixels that have been misclassified in the initial dense low-resolution prediction, but this is not feasible in practice. Inspired by recent works~\cite{huang2019uncertainty, abdar2021review} that utilize entropy maps to identify uncertain pixels, we adopt a similar thought and employ entropy as the criterion for selecting the pixels we need. Our intuition is that ``\emph{less confident predictions are more likely to be wrong}''.

The entropy selector uses the model's logits as input. Logits are the outputs of the segmentation model's final layer, produced just before applying the softmax operation. The size of the logits is $H\times W\times C$ where C is the number of classes. The selector calculates the entropy of each pixel using $e = -\sum_c p_c \log p_c$ (where $p$ $\in$ $R^C$). Pixels with high entropy (\ie exceeding a threshold $\alpha$) are selected. A PyTorch implementation is shown below:
\begin{minted}{python}
def entropy_selector(logits, threshold):
    probs = torch.softmax(logits, dim=-1)
    entropy = -torch.sum(probs * torch.log(probs), dim=-1)
    return entropy > threshold
\end{minted}
These selected pixels can then be extracted from the input image for further refinement.

\begin{figure}[t]
\centering
\includegraphics[width=\linewidth]{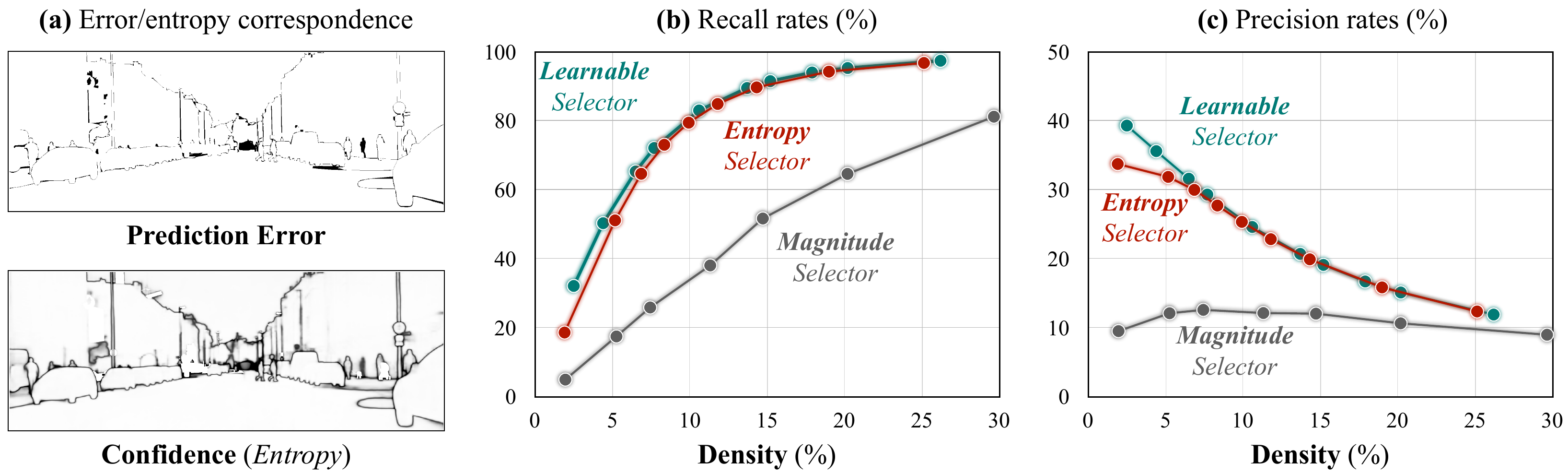}
\caption{The entropy map exhibits a strong correlation with the error map \textbf{(a)}. 
Recall rates \textbf{(b)} and precision rates \textbf{(c)} for the entropy selector, magnitude selector, and learnable selector.}
\label{fig:method:selector}
\vspace{-12pt}
\end{figure}

Visual verification from \fig{fig:method:selector}a confirms a strong correspondence between the entropy map and the error map. Quantitatively, our entropy selector is able to identify around 80\% of the misclassified pixels while selecting only 10\% of the total pixels (\fig{fig:method:selector}b). In contrast, the magnitude selector can only recover 40\% of them with a similar density. Although the learnable selector can achieve slightly higher recall rates than the entropy selector, it introduces higher latency, requiring 4.0ms on an NVIDIA RTX 3090 GPU. In contrast, our entropy selector is efficient, requiring only 2.3ms. Detailed comparison among the entropy selector, magnitude selector and the learnable selector is in Section~\ref{sec:analysis}. The precision (\fig{fig:method:selector}c) is less relevant in our case as the recall sets the accuracy upper bound for our refinement process.

\subsubsection{Sparse Feature Extractor.}

Having obtained a set of sparse pixels from the entropy selector, we then generate refinements for each. Processing sparse pixels presents unique challenges due to their irregular patterns compared to dense images. Interestingly, they share similarities with 3D point clouds, where occlusion is common and only geometric outlines are evident. Although sparse, the pixels retain a well-defined shape. The successful exploration of point cloud segmentation in previous works~\cite{wu2018squeezeseg,choy20194d,tang2020searching} provides valuable insight indicating that sparse pixels should also contain contextual information that can support our sparse refinement approach.

In this paper, we utilize a modified version of MinkowskiUNet~\cite{choy20194d} as our \textit{sparse feature extractor}. It follows the standard ResNet~\cite{he2016deep} basic block design with sparse convolutions and deconvolutions. Sparse convolution~\cite{graham20183d} is the sparse equivalent of conventional dense convolution with two main distinctions: firstly, sparse convolution avoids unnecessary computations for zero activations, and secondly, it preserves the same activation sparsity pattern throughout the model. These two properties make it much more efficient in processing our sparse pixel set. Furthermore, recent advances in system support for sparse convolution~\cite{yan2018second,choy20194d,tang2022torchsparse,hong2023pcengine,tang2023torchsparsev2} enable us to translate the theoretical computational reduction, resulting from sparsity, into actual measured speedup. Please note that though we have chosen sparse convolution in this paper, alternative designs are also feasible, such as point-based convolutions~\cite{qi2017pointnet++,li2018pointcnn,wang2018dynamic} and more recent point cloud transformers~\cite{fan2022embracing,sun2022swformer,liu2023flatformer,wang2023dsvt}.

The input to our sparse feature extractor is simply the raw RGB values of the selected pixels. We have explored adding more information from the low-resolution inference as input, such as final prediction or intermediate feature. While these additional features do contribute to faster convergence, they do not yield any improved performance. The output of our sparse feature extractor comprises multi-channel features for all selected pixels. We attach a simple linear classification head to generate the refinements for these pixels of interest.

\subsubsection{Gated Ensembler.}
\label{sec:method:gated_ensembler}
After obtaining the refinement predictions, the straightforward approach is to directly substitute the initial predictions at the corresponding pixels. However, this approach is not always optimal. This is because, compared to dense pixels, the context information available for sparse pixels in high-resolution is relatively limited. Incorporating initial predictions from dense low-resolution images, which provide more comprehensive context information, can be beneficial.

We introduce the \emph{gated ensembler} to intelligently combine the initial predictions ($y_1$) and the refined predictions ($y_2$). The key idea is to generate a weighting factor $w \in [0, 1]$ for each pixel of interest and utilize it to fuse the two predictions. Concretely, the final predictions are generated by
\begin{equation}
    y = f(w \cdot y_1 + (1 - w) \cdot y_2), \quad\text{where}~w = \text{sigmoid}(g([y_1; y_2; e_1; e_2])).
\end{equation}
Here, $f(\cdot)$ and $g(\cdot)$ are two-layer multi-layer perceptrons (MLPs). To generate the weighting factor, we provide both the raw predictions ($y_{1,2}$) and their corresponding entropies ($e_{1,2}$) as inputs to $g$.

\section{Experiments}
\label{sect:exp}

\begin{table*}[t]
% \caption{SparseRefine effectively closes the accuracy gap between low-resolution and high-resolution predictions, achieving a remarkable reduction in computational cost by \textbf{1.6 to 3.6 times} and inference latency by \textbf{1.5 to 3.9 times}. In this table, (\texttt{D}) and (\texttt{S}) denote dense and sparse inputs, respectively.}
\caption{SparseRefine effectively closes the accuracy gap between low-resolution and high-resolution predictions, achieving a remarkable reduction in computational cost by \textbf{1.4 to 3.1 times} and inference latency by \textbf{1.5 to 3.7 times}. In this table, (\texttt{D}) and (\texttt{S}) denote dense and sparse inputs, respectively.}
\setlength{\tabcolsep}{4pt}
\small\centering
\resizebox{\textwidth}{!}{
\begin{tabular}{lccccc}
\toprule
 & Input Resolution & \#Params (M) & \#MACs (T) & Latency (ms) & Mean IoU \\
\midrule
HRNet-W48 & 1024$\times$2048 (\texttt{D}) & 65.9 & \,\,0.75\,\,\tikzmark{hrnet-w48:a1} & \,\,53.4\,\,\tikzmark{hrnet-w48:a2} & \,\,80.7\,\,\tikzmark{hrnet-w48:a3} \\
\cdashlinelr{1-6}
\textcolor{gray!50}{HRNet-W48} & \textcolor{gray!50}{\,\,\,512$\times$1024 (\texttt{D})} & \textcolor{gray!50}{65.9} & \textcolor{gray!50}{0.19} & \textcolor{gray!50}{14.5} & \textcolor{gray!50}{79.2} \\
% ~+ \textbf{SparseRefine}  & 1024$\times$2048 (\texttt{S}) & 85.7 & \,\,{0.32}\,\,\tikzmark{hrnet-w48:b1} & \,\,{30.3}\,\,\tikzmark{hrnet-w48:b2} & \,\,{80.9}\,\,\tikzmark{hrnet-w48:b3} \\
~+ \textbf{SparseRefine}  & 1024$\times$2048 (\texttt{S}) & 145.2 & \,\,{0.38}\,\,\tikzmark{hrnet-w48:b1} & \,\,{32.4}\,\,\tikzmark{hrnet-w48:b2} & \,\,{80.9}\,\,\tikzmark{hrnet-w48:b3} \\
\midrule
\midrule
SegFormer-B5 & 1024$\times$2048 (\texttt{D}) & 82.0 & \,\,1.16\,\,\tikzmark{segformer-b5:a1} & \,\,140.6\,\,\tikzmark{segformer-b5:a2} & \,\,81.1\,\,\tikzmark{segformer-b5:a3} \\
% SegFormer-B5 & 640$\times$1280 (\texttt{D}) & 82.0 &  & 34.2 & 80.0 \\
\cdashlinelr{1-6}
\textcolor{gray!50}{SegFormer-B5} & \textcolor{gray!50}{\,\,\,512$\times$1024 (\texttt{D})} & \textcolor{gray!50}{82.0} & \textcolor{gray!50}{0.17} & \textcolor{gray!50}{18.5} & \textcolor{gray!50}{78.7} \\
% ~+ \textbf{SparseRefine}  & 1024$\times$2048 (\texttt{S}) & 101.8 & \,\,{0.32}\,\,\tikzmark{segformer-b5:b1} & \,\,{36.2}\,\,\tikzmark{segformer-b5:b2} & \,\,{81.2}\,\,\tikzmark{segformer-b5:b3} \\
~+ \textbf{SparseRefine}  & 1024$\times$2048 (\texttt{S}) & 161.3 & \,\,{0.38}\,\,\tikzmark{segformer-b5:b1} & \,\,{38.5}\,\,\tikzmark{segformer-b5:b2} & \,\,{81.2}\,\,\tikzmark{segformer-b5:b3} \\
\midrule
\midrule
Mask2Former-T & 1024$\times$2048 (\texttt{D}) & 36.7 & \,\,0.62\,\,\tikzmark{mask2former-t:a1} & \,\,66.8\,\,\tikzmark{mask2former-t:a2} & \,\,81.1\,\,\tikzmark{mask2former-t:a3} \\
\cdashlinelr{1-6}
\textcolor{gray!50}{Mask2Former-T} & \textcolor{gray!50}{\,\,\,512$\times$1024 (\texttt{D})} & \textcolor{gray!50}{36.7} & \textcolor{gray!50}{0.16} & \textcolor{gray!50}{19.1} & \textcolor{gray!50}{78.6} \\
% ~+ \textbf{SparseRefine}  & 1024$\times$2048 (\texttt{S}) & 56.5 & \,\,0.39\,\,\tikzmark{mask2former-t:b1} & \,\,{44.8}\,\,\tikzmark{mask2former-t:b2} & \,\,{81.3}\,\,\tikzmark{mask2former-t:b3} \\
~+ \textbf{SparseRefine}  & 1024$\times$2048 (\texttt{S}) & 56.5 & \,\,0.39\,\,\tikzmark{mask2former-t:b1} & \,\,{44.8}\,\,\tikzmark{mask2former-t:b2} & \,\,{81.1}\,\,\tikzmark{mask2former-t:b3} \\
\midrule
\midrule
Mask2Former-L & 1024$\times$2048 (\texttt{D}) & 207.0 & \,\,1.99\,\,\tikzmark{mask2former-l:a1} & \,\,150.8\,\,\tikzmark{mask2former-l:a2} & \,\,83.0\,\,\tikzmark{mask2former-l:a3} \\
\cdashlinelr{1-6}
\textcolor{gray!50}{Mask2Former-L} & \textcolor{gray!50}{\,\,\,512$\times$1024 (\texttt{D})} & \textcolor{gray!50}{207.0} & \textcolor{gray!50}{0.51} & \textcolor{gray!50}{45.4} & \textcolor{gray!50}{80.9} \\
% ~+ \textbf{SparseRefine}  & 1024$\times$2048 (\texttt{S}) & 226.8 & \,\,0.84\,\,\tikzmark{mask2former-l:b1} &  \,\,84.4\,\,\tikzmark{mask2former-l:b2} & \,\,83.0\,\,\tikzmark{mask2former-l:b3} \\
~+ \textbf{SparseRefine}  & 1024$\times$2048 (\texttt{S}) & 286.3 & \,\,0.93\,\,\tikzmark{mask2former-l:b1} &  \,\,89.9\,\,\tikzmark{mask2former-l:b2} & \,\,83.0\,\,\tikzmark{mask2former-l:b3} \\
\midrule
\midrule
SegNeXt-L & 1024$\times$2048 (\texttt{D}) & 48.8 & \,\,0.53\,\,\tikzmark{segnext-l:a1} & \,\,86.3\,\,\tikzmark{segnext-l:a2}  & \,\,83.0\,\,\tikzmark{segnext-l:a3} \\
\cdashlinelr{1-6}
\textcolor{gray!50}{SegNeXt-L} & \textcolor{gray!50}{\,\,\,640$\times$1280 (\texttt{D})} & \textcolor{gray!50}{48.8} & \textcolor{gray!50}{0.21} & \textcolor{gray!50}{33.6} & \textcolor{gray!50}{80.8} \\
% ~+ \textbf{SparseRefine}  & 1024$\times$2048 (\texttt{S}) & 68.6 & \,\,0.32\,\,\tikzmark{segnext-l:b1} & \,\,49.1\,\,\tikzmark{segnext-l:b2} & \,\,82.8\,\,\tikzmark{segnext-l:b3}\\
~+ \textbf{SparseRefine}  & 1024$\times$2048 (\texttt{S}) & 128.1 & \,\,0.37\,\,\tikzmark{segnext-l:b1} & \,\,50.9\,\,\tikzmark{segnext-l:b2} & \,\,82.8\,\,\tikzmark{segnext-l:b3}\\
\bottomrule
\end{tabular}

\begin{tikzpicture}[overlay, remember picture, shorten >=.5pt, shorten <=.5pt, transform canvas={yshift=.25\baselineskip}]
% \draw [->, darkred] ({pic cs:hrnet-w48:a1}) [bend left] to node [below right] (hrnet-w48:t1) {\hspace{-2pt}\scriptsize \textbf{2.3$\times$}} ({pic cs:hrnet-w48:b1});
\draw [->, darkred] ({pic cs:hrnet-w48:a1}) [bend left] to node [below right] (hrnet-w48:t1) {\hspace{-2pt}\scriptsize \textbf{2.0$\times$}} ({pic cs:hrnet-w48:b1});
% \draw [->, darkred] ({pic cs:hrnet-w48:a2}) [bend left] to node [below right] (hrnet-w48:t2) {\hspace{-2pt}\scriptsize \textbf{1.8$\times$}} ({pic cs:hrnet-w48:b2});
\draw [->, darkred] ({pic cs:hrnet-w48:a2}) [bend left] to node [below right] (hrnet-w48:t2) {\hspace{-2pt}\scriptsize \textbf{1.6$\times$}} ({pic cs:hrnet-w48:b2});
\draw [->] ({pic cs:hrnet-w48:a3}) [bend left] to node [below right] (hrnet-w48:t3) {\hspace{-2pt}\scriptsize +0.2} ({pic cs:hrnet-w48:b3});

% \draw [->, darkred] ({pic cs:segformer-b5:a1}) [bend left] to node [below right] (segformer-b5:t1) {\hspace{-2pt}\scriptsize \textbf{3.6$\times$}} ({pic cs:segformer-b5:b1});
\draw [->, darkred] ({pic cs:segformer-b5:a1}) [bend left] to node [below right] (segformer-b5:t1) {\hspace{-2pt}\scriptsize \textbf{3.1$\times$}} ({pic cs:segformer-b5:b1});
% \draw [->, darkred] ({pic cs:segformer-b5:a2}) [bend left] to node [below right] (segformer-b5:t2) {\hspace{-2pt}\scriptsize \textbf{3.9$\times$}} ({pic cs:segformer-b5:b2});
\draw [->, darkred] ({pic cs:segformer-b5:a2}) [bend left] to node [below right] (segformer-b5:t2) {\hspace{-2pt}\scriptsize \textbf{3.7$\times$}} ({pic cs:segformer-b5:b2});
\draw [->] ({pic cs:segformer-b5:a3}) [bend left] to node [below right] (segformer-b5:t3) {\hspace{-2pt}\scriptsize +0.1} ({pic cs:segformer-b5:b3});

\draw [->, darkred] ({pic cs:mask2former-t:a1}) [bend left] to node [below right] (mask2former-l:t1) {\hspace{-2pt}\scriptsize \textbf{1.6$\times$}} ({pic cs:mask2former-t:b1});
\draw [->, darkred] ({pic cs:mask2former-t:a2}) [bend left] to node [below right] (mask2former-t:t2) {\hspace{-2pt}\scriptsize \textbf{1.5$\times$}} ({pic cs:mask2former-t:b2});
% \draw [->] ({pic cs:mask2former-t:a3}) [bend left] to node [below right] (mask2former-l:t3) {\hspace{-2pt}\scriptsize +0.2} ({pic cs:mask2former-t:b3});
\draw [->] ({pic cs:mask2former-t:a3}) [bend left] to node [below right] (mask2former-l:t3) {\hspace{-2pt}\scriptsize +0.0} ({pic cs:mask2former-t:b3});

% \draw [->, darkred] ({pic cs:mask2former-l:a1}) [bend left] to node [below right] (mask2former-l:t1) {\hspace{-2pt}\scriptsize \textbf{2.4$\times$}} ({pic cs:mask2former-l:b1});
\draw [->, darkred] ({pic cs:mask2former-l:a1}) [bend left] to node [below right] (mask2former-l:t1) {\hspace{-2pt}\scriptsize \textbf{2.1$\times$}} ({pic cs:mask2former-l:b1});
% \draw [->, darkred] ({pic cs:mask2former-l:a2}) [bend left] to node [below right] (mask2former-l:t2) {\hspace{-2pt}\scriptsize \textbf{1.8$\times$}} ({pic cs:mask2former-l:b2});
\draw [->, darkred] ({pic cs:mask2former-l:a2}) [bend left] to node [below right] (mask2former-l:t2) {\hspace{-2pt}\scriptsize \textbf{1.7$\times$}} ({pic cs:mask2former-l:b2});
\draw [->] ({pic cs:mask2former-l:a3}) [bend left] to node [below right] (mask2former-l:t3) {\hspace{-2pt}\scriptsize +0.0} ({pic cs:mask2former-l:b3});

% \draw [->, darkred] ({pic cs:segnext-l:a1}) [bend left] to node [below right] (segnext-l:t1) {\hspace{-2pt}\scriptsize \textbf{1.7$\times$}} ({pic cs:segnext-l:b1});
\draw [->, darkred] ({pic cs:segnext-l:a1}) [bend left] to node [below right] (segnext-l:t1) {\hspace{-2pt}\scriptsize \textbf{1.4$\times$}} ({pic cs:segnext-l:b1});
% \draw [->, darkred] ({pic cs:segnext-l:a2}) [bend left] to node [below right] (segnext-l:t2) {\hspace{-2pt}\scriptsize \textbf{1.8$\times$}} ({pic cs:segnext-l:b2});
\draw [->, darkred] ({pic cs:segnext-l:a2}) [bend left] to node [below right] (segnext-l:t2) {\hspace{-2pt}\scriptsize \textbf{1.7$\times$}} ({pic cs:segnext-l:b2});
\draw [->] ({pic cs:segnext-l:a3}) [bend left] to node [below right] (segnext-l:t3) {\hspace{-2pt}\scriptsize --0.2} ({pic cs:segnext-l:b3});
\end{tikzpicture}}
% \vspace{4pt}
\label{tab:results:cityscapes}
% \vspace{-25pt}
\end{table*}
\begin{table*}[t]
\caption{SparseRefine generalizes across common object (Pascal VOC), autonomous driving (BDD100K), aerial (DeepGlobe) and medical (ISIC) datasets, achieving a \textbf{1.5-2.0$\times$} measured speedup with no loss of accuracy. The unit of latency is ms.}
\setlength{\tabcolsep}{4pt}
\small\centering
\resizebox{\textwidth}{!}{
\begin{tabular}{lccccccccc}
\toprule
 & & \multicolumn{2}{c}{Pascal VOC} & \multicolumn{2}{c}{BDD100K} & \multicolumn{2}{c}{DeepGlobe}  & \multicolumn{2}{c}{ISIC}  \\
%\cdashlinelr{1-8}
\midrule
 % & Resolution & Latency (ms) & mIoU & Latency (ms) & mIoU & Latency (ms) & mIoU \\
 & Resolution & Latency & mIoU & Latency & mIoU & Latency & mIoU & Latency & mIoU \\
\midrule
HRNet-W48 & Full (\texttt{D}) & \,\,14.7\,\,\tikzmark{bdd:a7} & \,\,77.8\,\,\tikzmark{bdd:a8} & \,\,23.5\,\,\tikzmark{bdd:a1} & \,\,63.6\,\,\tikzmark{bdd:a2} & \,\,146.4\,\,\tikzmark{bdd:a3} & \,\,73.4\,\,\tikzmark{bdd:a4} & \,\,157.7\,\,\tikzmark{bdd:a5} & \,\,82.3\,\,\tikzmark{bdd:a6} \\
\cdashlinelr{1-10}
\textcolor{gray!50}{HRNet-W48} & \textcolor{gray!50}{Half (\texttt{D})} & \textcolor{gray!50}{5.0} & \textcolor{gray!50}{77.2} & \textcolor{gray!50}{6.1} & \textcolor{gray!50}{60.7} & \textcolor{gray!50}{38.7} & \textcolor{gray!50}{72.9} & \textcolor{gray!50}{40.6} & \textcolor{gray!50}{80.8} \\
~+ \textbf{SparseRefine} & Full (\texttt{S}) &  \,\,8.1\,\,\tikzmark{bdd:b7} & \,\,{78.2}\,\,\tikzmark{bdd:b8}  &\,\,15.6\,\,\tikzmark{bdd:b1} & \,\,{63.5}\,\,\tikzmark{bdd:b2} & \,\,{92.9}\,\,\tikzmark{bdd:b3} & \,\,{73.4}\,\,\tikzmark{bdd:b4} & \,\,79.4\,\,\tikzmark{bdd:b5} & \,\,82.5\,\,\tikzmark{bdd:b6} \\
\bottomrule
\end{tabular}
% \begin{tikzpicture}[overlay, remember picture, shorten >=.5pt, shorten <=.5pt, transform canvas={yshift=.25\baselineskip}]
% \draw [->, darkred] ({pic cs:bdd:a1}) [bend left] to node [below right] (bdd:t1) {\hspace{-2pt}\scriptsize \textbf{1.5$\times$}} ({pic cs:bdd:b1});

% \draw [->] ({pic cs:bdd:a2}) [bend left] to node [below right] (bdd:t2) {\hspace{-2pt}\scriptsize -0.1} ({pic cs:bdd:b2});

% \draw [->, darkred] ({pic cs:bdd:a3}) [bend left] to node [below right] (bdd:t3) {\hspace{-2pt}\scriptsize \textbf{1.6$\times$}} ({pic cs:bdd:b3});

% \draw [->] ({pic cs:bdd:a4}) [bend left] to node [below right] (bdd:t4) {\hspace{-2pt}\scriptsize +0.0} ({pic cs:bdd:b4});

% \draw [->, darkred] ({pic cs:bdd:a5}) [bend left] to node [below right] (bdd:t5) {\hspace{-2pt}\scriptsize \textbf{2.0$\times$}} ({pic cs:bdd:b5});

% \draw [->] ({pic cs:bdd:a6}) [bend left] to node [below right] (bdd:t6) {\hspace{-2pt}\scriptsize +0.2} ({pic cs:bdd:b6});
% \end{tikzpicture}
}
\label{tab:results:additional}
\end{table*}

\subsection{Setup}

\subsubsection{Dataset.}

We evaluate SparseRefine primarily on Cityscapes~\cite{cordts2016cityscapes}, a dataset of 5,000 high-resolution (1024$\times$2048) urban scene images with pixel-level annotations for 19 semantic categories. To demonstrate generalizability, we validate our approach on four additional datasets: Pascal VOC~\cite{everingham2010pascal} for common objects, BDD100K~\cite{yu2020bdd100k} for autonomous driving, DeepGlobe~\cite{demir2018deepglobe} for aerial images, and ISIC~\cite{codella2018skin} for medical images. We employ mean Intersection-over-Union (mIoU) as the primary evaluation metric.

\subsubsection{Baselines.}

To showcase the generalizability of our method across diverse architectures, we employ five models spanning both convolutional and transformer-based approaches. We choose HRNet-W48~\cite{wang2020deep} as the convolution-based baseline, and SegFormer-B5~\cite{xie2021segformer}, Mask2Former-T~\cite{cheng2022masked}, Mask2Former-L~\cite{cheng2022masked}, and SegNeXt-L~\cite{guo2022segnext} as our transformer-based baselines. We reproduce the results of all high-resolution and low-resolution baselines using MMSegmentation v1.0.0~\cite{mmsegmentation}. We adhere to the default training settings, only making minimal adjustments to data augmentation parameters for lower resolutions. Due to observed instability in Mask2Former's results, we report the mean of three runs for a more robust evaluation.

\subsubsection{Model Details.}

As the prediction logits vary across different baseline model architectures, we trained separate sparse refinement module based on their respective low-resolution output logits to achieve the optimal results. We set a different entropy threshold for each baseline model in our entropy selector. Our sparse feature extractor is a modified MinkowskiUNet that has six stages with channel dimensions of 32, 64, 128, 256, 512 and 1024 for each stage by default. At each stage, there are two ResNet basic blocks before downsampling and another two after upsampling. Our gated ensembler employs two linear layers to produce the weighting factor and an additional two layers to combine the predictions, both with a hidden dimension of 64. We refer the readers to the appendix for more implementation details.

\subsubsection{Training Details.}

SparseRefine is trained independently from the dense baselines. We use the same data augmentation and training strategy employed by the dense baseline to ensure that performance improvements stem solely from our method. For data augmentation, we apply standard techniques such as random scaling (between 0.5 and 2.0), horizontal flipping, cropping (with a size of 512$\times$1024), and photometric distortion. We apply the standard cross entropy loss to supervise the model. We adopt AdamW~\cite{loshchilov2017decoupled} as our optimizer, with an initial learning rate of 0.0003 and a weight decay of 0.05. We gradually decay the learning rate following the cosine-annealing schedule~\cite{loshchilov2017sgdr}. We train the model for 500 epochs with a batch size of 32. The training takes around 12 hours on 8 NVIDIA RTX A6000 GPUs.

\subsubsection{Latency Details.}

We use cuBLAS~\cite{cublas} for all dense operations and utilize TorchSparse++~\cite{tang2022torchsparse,tang2023torchsparsev2} for all sparse operations. We measure the inference latency of all methods using a single NVIDIA RTX 3090 GPU with FP16 precision and a batch size of 4. We provide additional results for different precisions and batch sizes in the appendix. We omit batch normalization layers in latency measurement as they can be folded into preceding convolution layers. We report the average latency over 500 inference steps, with a 100-step warm-up period.

\subsection{Results}

\begin{figure}[t]
\centering
\includegraphics[width=\linewidth]{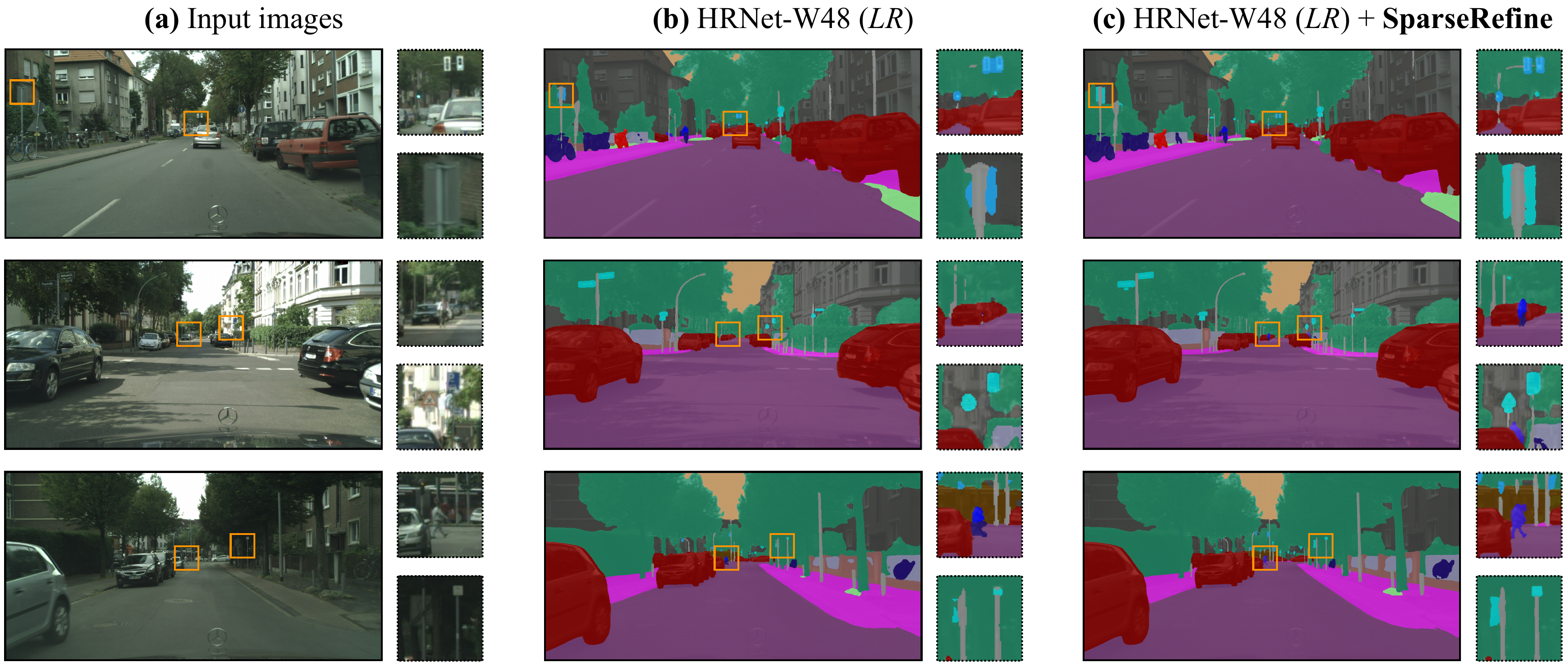}
\caption{SparseRefine improves the low-resolution (\textit{LR}) baseline with substantially better recognition of small, distant objects and finer detail around object boundaries.}
\label{fig:results:cityscapes}
\vspace{-10pt}
\end{figure}
\begin{table*}[t]
\caption{SparseRefine is better than token pruning and mask refinement approaches.}
\begin{subtable}[t]{0.48\textwidth}
\centering
\resizebox{\textwidth}{!}{
\setlength{\tabcolsep}{1pt}
\small
\begin{tabular}{lcc}
\toprule
& Latency (ms) & Mean IoU \\
\midrule    
Mask2Former-L & 150.8 & 83.0 \\
\midrule
~+ SparseViT & \,\,132.3\,\,\tikzmark{sparsevit:a} & 83.2 \\
~+ \textbf{SparseRefine} (Ours) & \,\,89.9\,\,\tikzmark{sparsevit:b} & 83.0 \\
\bottomrule
\end{tabular}
\begin{tikzpicture}[overlay, remember picture, shorten >=.5pt, shorten <=.5pt, transform canvas={yshift=.25\baselineskip}]
% \draw [->, darkred] ({pic cs:sparsevit:a}) [bend left] to node [right] (sparsevit:t) {\hspace{-1pt}\scriptsize \textbf{1.6$\times$}} ({pic cs:sparsevit:b});
\draw [->, darkred] ({pic cs:sparsevit:a}) [bend left] to node [right] (sparsevit:t) {\hspace{-1pt}\scriptsize \textbf{1.5$\times$}} ({pic cs:sparsevit:b});
\end{tikzpicture}}
\caption{Comparison to token pruning.}
\label{tab:comparison:a}
\end{subtable}
% \hfill
\hspace{1pt}
\begin{subtable}[t]{0.48\textwidth}
\centering
\resizebox{\textwidth}{!}{
\setlength{\tabcolsep}{1pt}
\small
\begin{tabular}{lcc}
\toprule
& Latency (ms) & Mean IoU \\
\midrule
HRNet-W48 & 53.4 & 80.7 \\
\midrule
% ~+ PointRend & 30.5 & \,\,79.8\,\,\tikzmark{pointrend:a} \\
~+ PointRend & 32.8 & \,\,79.9\,\,\tikzmark{pointrend:a} \\
% ~+ \textbf{SparseRefine} (Ours) & 30.3 & 80.9\tikzmark{pointrend:b} \\
~+ \textbf{SparseRefine} (Ours) & 32.4 & \,\,80.9\,\,\tikzmark{pointrend:b} \\
\bottomrule
\end{tabular}
\begin{tikzpicture}[overlay, remember picture, shorten >=.5pt, shorten <=.5pt, transform canvas={yshift=.25\baselineskip}]
% \draw [->, darkred] ({pic cs:pointrend:a}) [bend left] to node [right] (pointrend:t) {\hspace{-1pt}\scriptsize \textbf{+1.1}} ({pic cs:pointrend:b});
\draw [->, darkred] ({pic cs:pointrend:a}) [bend left] to node [right] (pointrend:t) {\hspace{-1pt}\scriptsize \textbf{+1.0}} ({pic cs:pointrend:b});
\end{tikzpicture}}
\caption{Comparison to mask refinement.}
\label{tab:comparison:b}
\end{subtable}
\label{tab:comparison}
\vspace{-20pt}
\end{table*}

We present our key experimental results in \tab{tab:results:cityscapes}. SparseRefine achieves substantial improvements in \#MACs and latency, while maintaining competitive or even better accuracy compared to the baselines. Specifically, SparseRefine accelerates the baselines by at least \textbf{1.5 times} and reduces the MACs by at least \textbf{1.4 times}. Notably, SparseRefine achieves a significant speedup of \textbf{3.7 times} for SegFormer-B5. This huge improvement can be attributed to the fact that SegFormer incorporates a vanilla self-attention module with high computational complexity ($O(H^2W^2)$), and our ``\textit{downsample then sparsely refine}'' strategy in SparseRefine can drastically reduce the computational cost via downsampling while recovering the accuracy through refining. From \tab{tab:results:additional}, SparseRefine demonstrates effective generalization across common objects(Pascal VOC), driving (BDD100K), aerial (DeepGlobe), and medical (ISIC) datasets. It delivers a consistent speedup from \textbf{1.5} to \textbf{2.0} times without compromising accuracy.%\zhuoyang{Add ADE and VOC results, both for the table and the text.}

In addition to the quantitative results, we also present qualitative results in \fig{fig:results:cityscapes}. In the middle column, it can be observed that the low-resolution baseline struggles to accurately classify pixels in distant areas and often misclassifies details near the edges. Our SparseRefine significantly improves the ambiguous predictions, as shown in the third column. The second row is a notable example, where the segmentation on low-resolution images fails to detect a person in the far distance, while SparseRefine accurately predicts their presence. Furthermore, SparseRefine even achieves accurate predictions in challenging cases, such as the thin rod of traffic lights in the first row. Quantitatively, we find that SparseRefine disproportionately improves the segmentation of smaller objects by almost 14$\times$ when compared to larger counterparts as discussed in the appendix. These results further demonstrate the effectiveness of our method.

\subsubsection{Comparison to Token Pruning.}

SparseViT~\cite{chen2023sparsevit} is a pioneering work that demonstrates the viability of token pruning for dense prediction tasks like semantic segmentation. As in \tab{tab:comparison:a}, SparseRefine exhibits a notable advantage over SparseViT in terms of latency reduction, achieving a 1.5$\times$ speedup while maintaining comparable accuracy. SparseRefine operates independently from token pruning methods, making it potentially compatible for use alongside them.
\vspace{-20pt}
\subsubsection{Comparison to Mask Refinement.}

We also compare SparseRefine with PointRend~\cite{kirillov2020pointrend}, a mask refinement method. To ensure fair comparisons, we adjust the images to a resolution of 704$\times$1408 for PointRend, aligning its latency with our SparseRefine. As in \tab{tab:comparison:b}, PointRend suffers a performance decline, while our method demonstrates an improvement over the baseline. PointRend relies on an MLP-based mask refinement approach using hidden features. However, when low-resolution images are used as input, the refinement process struggles to effectively compensate for information loss caused by downsampling. SparseRefine, by working directly on the high-resolution image, minimizes information loss and consequently boosts performance.

\subsubsection{Comparison to Patch Refinement.}

\begin{wraptable}{r}{0.5\textwidth}
\vspace{-30pt}
\caption{Sparse refinement is faster and more accurate than patch refinement.}
\vspace{4pt}
\setlength{\tabcolsep}{2pt}
\small\centering
\begin{tabular}{ccc}
\toprule
 & Latency (ms) & mIoU \\
\midrule 
Baseline & 53.4 & 80.7 \\
\midrule
PatchRefine (512) & 44.1 & 80.8 \\
PatchRefine (256) & 55.4 & 80.8 \\
\midrule
SparseRefine & \textbf{32.4} & \textbf{80.9} \\
\bottomrule
\end{tabular}
\vspace{-10pt}
\label{tab:results:patch}
\end{wraptable}

Some existing methods~\cite{huang2019uncertainty,wu2020patch,verelst2022segblocks} refine the prediction in a \textit{coarse-grained patch} level, while SparseRefine refines the prediction in a \textit{fine-grained pixel} level. As depicted in \fig{fig:method:selector}a, errors tend to be \textit{scattered sparsely} across the entire image, making fine-grained sparsity a more suitable solution. Patch-based refinement can often lead to substantial redundant computation, as not every pixel within a patch may need refinement. This inefficiency renders the patch-based methods less effective. From \tab{tab:results:patch}, SparseRefine outperforms patch refinement baselines (with a patch size of 256 or 512), achieving a speedup of \textbf{1.4} to \textbf{1.7} times while also delivering higher accuracy.
\subsection{Analysis}
\label{sec:analysis}

We analyze various alternative designs for the components of our method. We also provide detailed breakdowns of the improvements in both accuracy and efficiency. We use HRNet-W48 as the baseline model for analyses in this section.

\begin{table*}[t]
\caption{Ablation experiments to validate our design choices. Default settings are marked in  \colorbox{blue!10}{blue}.}
% \renewcommand\arraystretch{1.1}
% \centering
\begin{subtable}[t]{0.54\textwidth}
\setlength{\tabcolsep}{2pt}
\small\centering
\begin{tabular}{ccccc}
\toprule
% Criteria & Density & Precision & Recall & mIoU \\
Criteria & Density & Recall & Latency & mIoU\\
\midrule 
% Random & 10.0\% & 0.3\% & 10.0\% & 79.2\\
Random & 10.0\% & 10.0\% & - & 79.2\\
% Magnitude & 11.3\% & 12.1\% & 38.1\% & 80.2\\
Magnitude & 11.3\% & 38.1\% & 0.5ms & 80.2\\
% Learnable &  11.8\% & 91.2\% & 17.2ms & 81.1\\
% Learnable &  11.8\% & 85.4\% & 5.3ms & 80.9\\
Learnable &  11.8\% & 85.9\% & 4.0ms & 80.9\\
% \rowcolor{blue!10}Entropy & 11.8\% & \textbf{22.8\%} & \textbf{84.9\%} & \textbf{80.9} \\
% \rowcolor{blue!10}Entropy & 11.8\%  & 84.9\% & 2.0ms & 80.9\\
\rowcolor{blue!10}Entropy & 11.8\%  & 84.9\% & 2.3ms & 80.9\\
\textcolor{gray!50}{Oracle} & \textcolor{gray!50}{3.3\%} &
\textcolor{gray!50}{100\%} & 
- &
\textcolor{gray!50}{92.8} \\
\bottomrule
\end{tabular}
\caption{\textbf{Pixel Selector.} Entropy is an effective and efficient indicator for identifying misclassified pixels.
%\zhuoyang{Fill} 
} 
\label{tab:ablation:a}
\end{subtable}
\hspace{4pt}
% \hfill
\begin{subtable}[t]{0.42\textwidth}
\setlength{\tabcolsep}{3pt}
\small\centering
\begin{tabular}{ccccc}
\toprule
$\alpha$ & Density & Recall & Latency & mIoU \\
\midrule
% 0.8 & 3.4\%  & 32.2\%  &  23.1 ms & 79.9\\
0.8 & 3.4\%  & 32.2\%  &  24.6 ms & 79.9\\
% 0.6 & 6.9\%  & 64.6\%  &  27.0 ms   & 80.4 \\
0.6 & 6.9\%  & 64.6\%  &  28.9 ms   & 80.4 \\
% \rowcolor{blue!10}0.3 & 11.8\% & 84.9\%  &  30.3 ms & 80.9\\
\rowcolor{blue!10}0.3 & 11.8\% & 84.9\%  &  32.4 ms & 80.9\\
% 0.1 & 19.0\% & 94.3\%  &  37.3 ms & 81.1 \\
0.1 & 19.0\% & 94.3\%  &  40.1 ms & 81.1 \\
-- & 100.0\% & 100.0\%  &  -- &  80.5 \\
\bottomrule
\end{tabular}
\caption{\textbf{Entropy Threshold.} Performance improves with more pixels kept, but latency also increases.} 
\label{tab:ablation:b}
\end{subtable}

% \vspace{8pt}

\begin{subtable}[t]{0.3\textwidth}
\setlength{\tabcolsep}{1pt}
\small\centering
\begin{tabular}{cccc}
\toprule
RGB & Logits & Features & mIoU \\
\midrule
\rowcolor{blue!10}\cmark & \xmark & \xmark & 80.9 \\   
% \xmark & \cmark & \xmark & 79.7 \\
\cmark & \cmark & \xmark & 80.2\\
\cmark & \xmark & \cmark &  80.4 \\
\cmark & \cmark & \cmark & 80.4 \\
\bottomrule
\end{tabular}
\caption{\textbf{Input of sparse feature extractor}. Raw RGB provides enough information.} 
\label{tab:ablation:c}
\end{subtable}
\hspace{3pt}
\begin{subtable}[t]{0.46\textwidth}
\setlength{\tabcolsep}{1pt}
\small\centering
\begin{tabular}{cccc}
\toprule
 & \# of Channels (32$\times$) & mIoU \\
\midrule
MinkUNet & \{1,2,4,8\} & 80.5 \\
% \rowcolor{blue!10}MinkUNet & \{1,2,4,8,16\} & 80.9 \\
\rowcolor{blue!10}MinkUNet & \{1,2,4,8,16,32\} & 80.9 \\
MinkUNet & \{1,2,4,8,12,16,24,32\} & 80.9 \\
PointNet & -- & 79.3 \\
\bottomrule
\end{tabular}
\caption{\textbf{Architecture of sparse feature extractor}. MinkUNet is much better than PointNet.} 
\label{tab:ablation:d}
\end{subtable}
\hspace{3pt}
\begin{subtable}[t]{0.18\textwidth}
\setlength{\tabcolsep}{1pt}
\small\centering
\begin{tabular}{cc}
\toprule
Strategy & mIoU \\
\midrule
Direct & 77.7 \\   
Entropy & 80.3 \\
\rowcolor{blue!10}Gated & 80.9 \\
\textcolor{gray!50}{Oracle} & \textcolor{gray!50}{85.3} \\
\bottomrule
\end{tabular}
\caption{\textbf{Ensembler.} Gated ensembler is the best.} 
\label{tab:ablation:e}
\end{subtable}
% \vspace{-8pt}
\label{tab:ablation}
\vspace{-15pt}
\end{table*}
\begin{table*}[t]
\centering
\begin{minipage}{0.5\textwidth}
\caption{\textbf{Breakdown of \#MACs and latency}. Entropy selector and gated ensembler are lightweight.}
\small\centering
\begin{tabular}{lcc}
\toprule
 & \#MACs & Latency \\
\midrule
Entropy Selector  & 0 & 2.3 ms \\
% Sparse Feature Extractor & 0.129T  & 10.9 ms \\
Sparse Feature Extractor & 0.184T  & 13.0 ms \\
Gated Ensembler &  0.001T & 2.2 ms \\
\bottomrule
\end{tabular}
\label{tab:analysis:breakdown}
\end{minipage}
\hspace{4pt}
\begin{minipage}{0.46\textwidth}
\caption{\textbf{Sparse inference backend}. Sparse inference is more efficient than dense inference.} 
\small\centering
\begin{tabular}{lcc}
\toprule
Backend & Activation & Latency \\
\midrule
% cuBLAS & Dense & 52.0 ms \\   
cuBLAS & Dense & 62.3 ms \\   
% SpConv v2.3.5 & Sparse & 14.2 ms \\
SpConv v2.3.5 & Sparse & 15.8 ms \\
% TorchSparse v2.1.0 & Sparse & \textbf{10.9 ms} \\
TorchSparse v2.1.0 & Sparse & \textbf{13.0 ms} \\
\bottomrule
\end{tabular}
\label{tab:analysis:backend}
\end{minipage}
\vspace{-12pt}
\end{table*}

\subsubsection{Pixel Selector.}

We compare our proposed entropy-based pixel selector to other alternatives, including random selector, magnitude selector, and learnable selector as shown in \tab{tab:ablation}a. The random selector randomly selects pixels based on a density hyperparameter that we set. Compared to the entropy-based selector, the random selector shows a substantial performance drop of 1.7 mIoU, failing to achieve any improvement over the low-resolution baseline. This decline can be attributed to the notably low recall rate (10\%) of the random selection approach and its lack of principled criteria to ensure the selection of misclassified pixels.

Another alternative is the magnitude-based selector. It calculates the L2 magnitude on the output of the last layer, just before the segmentation head, in order to obtain an importance score for each pixel. This approach is commonly employed in token pruning works to identify and remove unimportant areas. As depicted in \tab{tab:ablation}a, it is evident that the magnitude-based selector still exhibits significantly lower recall and precision compared to the entropy-based selector. Consequently, the magnitude-based selector performs worse than our entropy selector by 0.7 mIoU.

The other alternative is the learnable selector. We first apply two dense convolutions on the RGB image to obtain a feature representation. Next, we combine the features with the dense baseline logits and entropy, sending the concatenated input through a multi-layer perceptron. We supervise the training of this selector independent of the SparseRefine pipeline. As shown in \Cref{tab:ablation}a, the learnable selector achieves a better recall when compared to our entropy selector, however it incurs almost $2\times$ the latency cost. Compared with the learnable selector, the entropy selector is much more efficient.

Furthermore, we also showcase the performance of the oracle setting, wherein we select incorrect predictions solely based on the ground truth. This highlights the considerable room for improvement and underscores the immense potential of our proposed paradigm.

\subsubsection{Entropy Threshold.} 

We present an analysis of the impact of different entropy thresholds on latency and accuracy, as outlined in \tab{tab:ablation}b. In essence, the entropy threshold involves a trade-off between latency and accuracy: a lower entropy threshold leads to the selection and refinement of more pixels, resulting in improved performance but increased latency. We select a moderate setting with $\alpha = 0.3$ for HRNet-W48, which matches the accuracy of the high-resolution baseline with the largest speedup. The optimal $\alpha$ for different models could be different. We provide more details in the appendix. We also investigate incorporating all the pixels to conduct ``dense'' refinement and obtain a result of 80.5, lower than 80.9 we got with 11.8\% density. This is reasonable because our goal is to refine the pixels that are ``difficult to learn in the low-resolutio'' ones. However, incorporating all the pixels would also involve including many pixels that are easy to learn. This can serve as a shortcut for the model to easily achieve low loss with weaker capability.

\subsubsection{Input of Sparse Feature Extractor.} 

We examine the use of various inputs for our sparse feature extractor, as illustrated in \tab{tab:ablation}c. Our default setting is to directly feed RGB pixels into the sparse feature extractor. In comparison to other settings that introduce additional logits and features, using purely RGB as input offers greater flexibility and underscores the advantage of SparseRefine as a plug-and-play module. Furthermore, we observe that incorporating more features in the inputs does not enhance performance. This suggests that the presence of ambiguous or even erroneous logits and features corresponding to misclassified pixels may mislead the sparse feature extractor, ultimately hindering performance improvement.

\subsubsection{Architecture of Sparse Feature Extractor.}

We analyze how different model capacities impact performance, as demonstrated in \ref{tab:ablation}d. Specifically, we incrementally increase the capacity of MinkUNet by expanding channels and adding more stages. The results indicate that performance improves as the model becomes larger, but eventually reaches saturation at 80.9 mIoU. We hypothesize that this occurs because we do not select all low-confidence pixels, which hinders further improvement in larger models. Additionally, we explore using PointNet as the sparse feature extractor. Similar to PointRend, its performance is also limited. The subpar performance may be attributed to the challenges associated with handling per-point RGB values without considering the contextual information.

\subsubsection{Ensembler.}

We investigate different ensemble strategies in Table \ref{tab:ablation}e. The simplest approach is to directly replace the initial predictions with the refined predictions. However, this is suboptimal due to SparseRefine's limited context (discussed in \sect{sec:method:gated_ensembler}). Another alternative is the entropy-based ensembler that compares the entropy before and after refinement to determine which predictions to choose. In comparison, our gated ensembler offers a softer and more compact way to incorporate refinement into the prediction. It is noteworthy that the gated ensembler outperforms the entropy-based ensembler by 0.6 mIoU. Additionally, we analyze the performance of the oracle setting, where we choose the better of the predictions before and after refinement. This analysis reveals a substantial room for potential improvement of 4.4 mIoU.

\subsubsection{Breakdowns.}

Our per-class performance in Table \ref{table: per class iou} reveals that SparseRefine consistently improves the performance of the low-resolution baseline in almost every category, particularly for small instances such as person, rider, pole. These categories also exhibit the most significant degradation in the low-resolution baseline when compared to the high-resolution baseline. This observation highlights the effectiveness of SparseRefine in capturing fine-grained details, thanks to its ability to utilize sparse high-resolution information.

\begin{table}[t]
\caption{SparseRefine consistently improves the performance of the low-resolution baseline across different categories, particularly for small objects.}
\setlength{\tabcolsep}{2pt}
\small\centering
\resizebox{1\textwidth}{!}{
\begin{tabular}{lcccccccccccccccccccc}
\toprule
 &  \rotatebox{90}{Road} & \rotatebox{90}{Sidewalk} & \rotatebox{90}{Building} & \rotatebox{90}{Wall} & \rotatebox{90}{Fence} & \rotatebox{90}{Pole} & \rotatebox{90}{Traffic Light} & \rotatebox{90}{Traffic Sign} & \rotatebox{90}{Vegetation} & \rotatebox{90}{Terrain} & \rotatebox{90}{Sky} & \rotatebox{90}{Person} & \rotatebox{90}{Rider} & \rotatebox{90}{Car} & \rotatebox{90}{Truck} & \rotatebox{90}{Bus} & \rotatebox{90}{Train} & \rotatebox{90}{Motorcycle} & \rotatebox{90}{Bicycle} & \rotatebox{90}{mIoU} \\
\midrule
HRNet-W48 (512$\times$1024) & 98.3 & 86.1 & 92.8 & \textbf{57.8} & \textbf{66.8} & 65.7 & 70.0 & 78.9 & 92.4 & \textbf{64.7} & 94.8 & 81.1 & 62.3 & 95.0 & 84.3 & 88.9 & 82.6 & 64.8 & 77.4 & 79.2\\
+ \textbf{SparseRefine} (Ours) & \textbf{98.4} & \textbf{86.7} & \textbf{93.4} & 57.7 & 66.7 & \textbf{70.5} & \textbf{75.0} & \textbf{82.0} & \textbf{93.0} & 64.4 & \textbf{95.4} & \textbf{84.3} & \textbf{67.4} & \textbf{95.8} & \textbf{85.6} & \textbf{89.6} & \textbf{83.3} & \textbf{67.9} & \textbf{80.1} & \textbf{80.9} \\
\midrule
\textcolor{gray!50}{HRNet-W48 (1024$\times$2048)} & \textcolor{gray!50}{98.4} & \textcolor{gray!50}{86.6} & \textcolor{gray!50}{93.2} & \textcolor{gray!50}{55.7} & \textcolor{gray!50}{64.9} & \textcolor{gray!50}{71.5} & \textcolor{gray!50}{75.8} & \textcolor{gray!50}{82.9} & \textcolor{gray!50}{92.8} & \textcolor{gray!50}{65.4} & \textcolor{gray!50}{95.4} & \textcolor{gray!50}{84.6} & \textcolor{gray!50}{65.8} & \textcolor{gray!50}{95.7} & \textcolor{gray!50}{80.4} & \textcolor{gray!50}{91.5} & \textcolor{gray!50}{83.2} & \textcolor{gray!50}{70.1} & \textcolor{gray!50}{80.1} & \textcolor{gray!50}{80.7} \\
\bottomrule
\end{tabular}}
\label{table: per class iou}
\vspace{-15pt}
\end{table}

\#MACs and latency breakdown for each component is presented in Table \ref{tab:analysis:breakdown}. The entropy selector and gated ensembler introduce minimal computational overhead, with the feature extractor remaining the primary computational component.  We have implemented the sparse feature extractor using different inference backends. As shown in Table \ref{tab:analysis:backend}, our input activation has high (approximately 90\%) sparsity. Therefore, sparse inference backends like SpConv and TorchSparse are more suitable than dense inference backends such as cuBLAS.
\section{Conclusion}

We present SparseRefine that enhances low-resolution dense predictions with high-resolution sparse refinements. It first incorporates an entropy selector to identify a sparse set of pixels with the lowest confidence, followed by a sparse feature extractor that efficiently generates refinements for those selected pixels. Finally, a gated ensembler is utilized to integrate these sparse refinements with the initial coarse predictions. Notably, SparseRefine can be seamlessly integrated into various existing semantic segmentation models, irrespective of their model architectures. Empirical evaluation on the five dataset demonstrated remarkable speed improvements, with negligible to no loss of accuracy. We believe that the speedups that accrue from our approach of combining \textit{low-resolution} prediction followed by \textit{sparse high-resolution} refinement, will further enable the deployment of high-resolution semantic segmentation in latency-sensitive applications. 

\myparagraph{Acknowledgement.}

This work was supported by MIT-IBM Watson AI Lab, MIT AI Hardware Program, and National Science Foundation.

\bibliographystyle{splncs04}
\bibliography{reference.bib}

\end{document}